\documentclass[10pt,twocolumn,letterpaper]{article}

\usepackage{cvpr}
\usepackage{times}
\usepackage{epsfig}
\usepackage{graphicx}
\usepackage{amsmath}
\usepackage{amssymb}
\usepackage{times}
\usepackage{multirow}
\usepackage{subfig}
\usepackage{amsthm}
\usepackage{bbm}
\usepackage[ruled,vlined]{algorithm2e}
\usepackage{makecell}
\usepackage{rotating}
\usepackage{booktabs}
\usepackage{float}
\usepackage{color}
\usepackage{bm}
\usepackage{subfloat}
\usepackage{threeparttable}

\newcommand{\vect}[1]{\mathbf{#1}}
\newcommand{\matr}[1]{\mathbf{#1}}

\newcommand{\set}[1]{\mathbb{#1}}

\usepackage[pagebackref=true,breaklinks=true,letterpaper=true,colorlinks,bookmarks=false]{hyperref}

 \cvprfinalcopy % *** Uncomment this line for the final submission

\def\cvprPaperID{3865} % *** Enter the CVPR Paper ID here

\ifcvprfinal\fi
\begin{document}

%%%%%%%%% TITLE
\title{Learning for Multi-Model and Multi-Type Fitting}

\author{Xun Xu\\
National University of Singapore\\
{\tt\small elexuxu@nus.edu.sg}
\and
Loong Fah Cheong\\
National University of Singapore\\
{\tt\small eleclf@nus.edu.sg}
\and
Zhuwen Li\\
Intel Labs\\
{\tt\small li.zhuwen@intel.com}
}

\maketitle
%\thispagestyle{empty}

%%%%%%%%% ABSTRACT
\begin{abstract}
  Multi-model fitting has been extensively studied from the random sampling and clustering perspectives. Most assume that only a single type/class of model is present and their generalizations to fitting multiple types of models/structures simultaneously are non-trivial. The inherent challenges include choice of types and numbers of models, sampling imbalance and parameter tuning, all of which render conventional approaches ineffective. In this work, we formulate the multi-model multi-type fitting problem as one of learning deep feature embedding that is clustering-friendly. In other words, points of the same clusters are embedded closer together through the network. For inference, we apply K-means to cluster the data in the embedded feature space and model selection is enabled by analyzing the K-means residuals. Experiments are carried out on both synthetic and real world multi-type fitting datasets, producing state-of-the-art results. %Not only is the network capable of dealing with the aforementioned issues, it is also able to disambiguate cases which involve criteria that are hard to specify analytically.
   Comparisons are also made on single-type multi-model fitting tasks with promising results as well.
\end{abstract}

%%%%%%%%% BODY TEXT
\section{Introduction}

Multi-model fitting has been a key problem in computer vision for decades. It aims to discover multiple independent structures, e.g. lines, circles, rigid motions, etc, often in the presence of noise. Here, by multi-model, we mean there are multiple models of a specific type, e.g. lines only. If in addition, there is a mixture of types (e.g. both lines and circles), we specifically term the problem as multi-model multi-type.

Various attempts towards solving the multi-model clustering problem have been made. %One class of approach is based on the hypothesis-and-test paradigm. 
The early works tend to be based on extensions of RANSAC \cite{fischler1981random} to the multi-model setting, e.g. simply running RANSAC multiple times consecutively \cite{torr1998geometric,vincent2001detecting}. More recent works in this approach involve analyzing the interplay between data and hypotheses. J-Linkage \cite{toldo2008robust}, its variant T-Linkage \cite{Magri2014} and ORK \cite{Chin2009,chin2010accelerated} rely on extensively sampling hypothesis models and compute the residual of data to each hypothesis. Either clustering is carried out on the mapping induced by the residuals, or an energy minimization is performed on the point to model distance, and various regularization terms (e.g. the label count penalty \cite{li2007two} and spatial smoothness (PEaRL) \cite{Isack2012}). Another class of approach involves direct analytic expressions characterizing the underlying subspaces, e.g., the powerful self-expressiveness assumption has inspired various elegant methods \cite{Elhamifar2013,liu2013robust,Li2015,Ji2016}. %. Here, there is often no need to make explicit the underlying model, e.g., the powerful self-expressiveness assumption has inspired various elegant methods \cite{Elhamifar2013,liu2013robust,Li2015,Ji2016}. %Along with the refinement of models, the performances on existing datasets have been pushed to the limits. 

\begin{figure}[!t]
\centering
\includegraphics[width=0.9\linewidth]{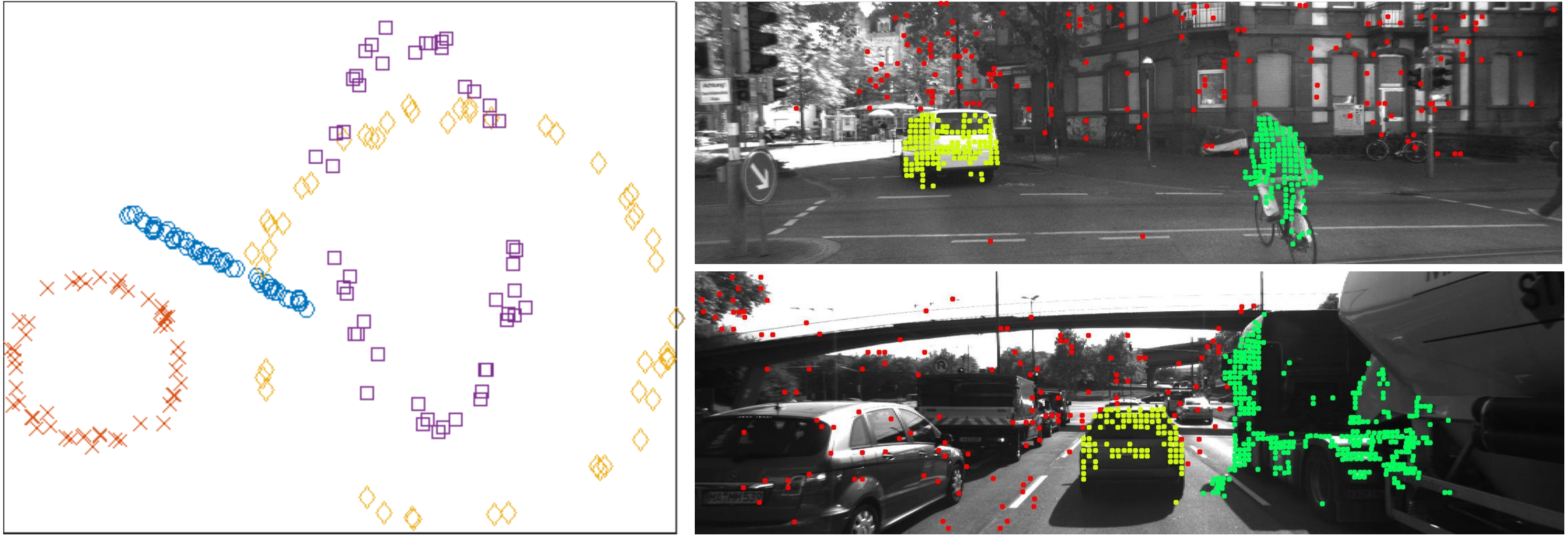}
\caption{Multi-model and multi-type fitting examples. Left: Simultaneously separating line, circle and ellipses. Right: 3D motion segmentation where motion can be explained by affine transform, homography and fundamental matrix on the right.}\label{fig:MultiTypeIllust}
\vspace{-0.5cm}
\end{figure}

Despite the considerable development of multi-model fitting techniques in the past two decades, there are still major lacuna in the problem. First of all, in contrast with having multiple instances of the same type/class, many real world model fitting problem consists of data sampled from multiple types of models. Fig.~\ref{fig:MultiTypeIllust} shows both a toy example of line, circle and ellipses co-existing together, and a realistic motion segmentation scenario, where the appropriate model to fit the foreground object motions (or even the background) can waver between affine motions, homography, and fundamental matrix \cite{xu2018motion} with no clear division. With few exceptions \cite{Barath2018,sugaya2004geometric,torr1998geometric}, none of the aforementioned works have considered this realistic scenario. Even if one attempts to fit multiple types of model sequentially like in \cite{sugaya2004geometric}, it is non-trivial to decide the type when the dichotomy of the models is unclear in the first place. Secondly, for problems where there are a significant number of models, the hypothesis-and-test approach is often overwhelmed by sampling imbalance, i.e., points from the same subspace represent only a minority, rendering the probability of hitting upon the correct hypothesis very small. This problem becomes severe when a large number of data samples are required for hypothesizing a model (e.g., eight points are needed for a linear estimation of the fundamental matrix and 5 points for fitting an ellipse). Lastly, for optimal performance, there is inevitably a lot of manipulation of parameters needed, among which the most sensitive include those for deciding what constitutes an inlier for a model \cite{Magri2014, Magri2015}, for sparsifying the affinity matrices \cite{Lai2017,xu2018motion}, and for selecting the model type \cite{torr1998geometric}. Often, dataset-specific tuning is required, with very little theory to guide the tuning.

There has been some recent foray into deep learning as a means to learn geometric model, e.g. camera pose \cite{brachmann2017dsac} and essential matrix \cite{yi2018learning} from feature correspondences, but extending such deep geometric model fitting approach to the multi-model and multi-type scenario has not been attempted. Generalizing the deep learning counterparts of RANSAC to multi-model fitting is not trivial due to the same reason as conventional sequential approaches. %Furthermore, in many geometric model fitting problems, the inliers belonging to a model instance are not necessarily clustered together in the input domain, so when multiple instances are present, one cannot detect them by some kind of window-based scanning mechanism. Instead, the deep network should be able to leverage on any potentially hidden regularities to discern the models despite the overwhelming sample imbalance. 
Furthermore, in many geometric model fitting problems, there are often significant overlap between the subspaces occupied by the multiple model instances (e.g. in motion segmentation, both the foreground and the background contain the camera-induced motion). We want the network to learn the best representation so that the different model instances can be well-separated. This is in contrast to the traditional clustering approaches where hand-crafted design of the similarity metric %or assumption of the kernel function 
is needed.%, and as a result these approaches suffer from the problems discussed before. 
When there are no clear division between multiple types of  models (e.g. the transitions from a circle to an ellipse), the network would also need to learn the appropriate preference from the labelled examples in the training data. %, without being given any explicit bias such as the various information criteria (AIC, BIC or MDL).

Another open
challenge in multi-model fitting is to automatically determine
the number of models, also referred to as model selection
in the literature \cite{Tibshirani2001,Chin2009,Li_2013_ICCV,Lai2017}. Traditional methods proceed from statistical analysis of the residual of the clustering \cite{Tibshirani2001,rousseeuw1987silhouettes}. %The well-known spectral clustering\cite{von2007tutorial} has theoretical analysis into selecting the number of clusters, however in a noise-free situation. 
Other methods approach from various heuristic standpoints including analyzing eigen values \cite{zelnik2005self,von2007tutorial}, over-segment and merge \cite{Li_2013_ICCV,Lai2017}, soft thresholding \cite{liu2013robust} or adding penalty terms \cite{Li2014}. %Most efforts in these work are dedicated to inferring correct number of clusters from noisy feature representations/affinities. It becomes more difficult combining multi-types of models, which has rarely if not none touched by the community. 
Most of the above works cannot deal with mixed-types in the models. To redress this gap in the literature, we want our network to learn good feature representations so that the number of clusters, even in the presence of mixed types, can be readily estimated.

With the above objectives in mind, we propose a multi-model multi-type fitting network. The network is given labelled data (inlier points for each model and outliers) and is supposed to learn the various geometric models in a completely data-driven manner. Since the input to the network is often not regular grid data like images, we use what we called the CorresNet from \cite{yi2018learning} as a backbone (see Fig.~\ref{fig:Network}). As the output of network should be amenable  for grouping into the respective, possibly mixed models, and invariant to any permutation of model indices among the multiple instances of the same class in the training data, we consider both an existing metric learning loss and its variant and propose a new distribution aware loss, the latter based on Fisher linear discriminant analysis (LDA). In the testing phase, standard K-means clustering is applied to the feature embeddings to obtain a discrete cluster assignment. As feature points are embedded in a clustering friendly way, we can just look into the K-means fitting residual to estimate the number of models should it be unknown.
%  Thanks to the ability of learning good feature representations, we can even robustly estimate the number of clusters in multi-type fitting by looking into the K-means residuals.
%, such simple approach produces results better than the state-of-the-art methods in multi-type fitting tasks.
%   knowledge about the underlying geometric models can be used to further refine the results. For this purpose, we propose a conservative RANSAC refinement step, whereby we suspend judgement about what constitutes as noise, since we cannot confidently set a global threshold that works well for all data. That is, if a data point is found not to meet the inlier threshold of any of the models found by RANSAC, we retain the original membership as output by the learnt network.  

%Very few works have tackled the multi-
%
%
%New solution from sparse data deep learning
%PointNet \cite{Qi2017}, PointNet++ \cite{QiLSG_NIPS2017}
%
%Learn geometric model from noisy data
%\cite{yi2018learning}.
%
%Existing pointnet can not deal with multiple instances. Therefore, we introduce metric learning to learn non-linear embedding according to the labels.
%
%Metric learning
%Affinity matrix regression \cite{hershey2016deep}, semi-hard triplet loss \cite{Schroff2015}, N-pair loss \cite{Sohn2016}, \cite{Song2017}
%
%Contributions:
%We propose to exploit sparse data deep network with metric learning loss. 
%Explicit encode geometric model prior as fitting residual

\section{Related Work}

\noindent\textbf{Multi-Model Fitting}:
Early approaches address this in a sequential RANSAC fashion \cite{torr1998geometric,vincent2001detecting,Yasushi2004}. These algorithms iteratively fit one model per iteration using RANSAC and remove the inliers of the best model. The J-Linkage \cite{toldo2008robust} and its variant T-Linkage \cite{Magri2014} simultaneously consider the interactions between all points and hypotheses. The final partition is achieved in an agglomerative clustering manner. The above greedy algorithms often do not perform well under high noise level. In contrast, global algorithms have been proposed to minimize an energy with various regularization terms. These include spatial regularization (PEaRL) \cite{Isack2012} and label count penalty \cite{li2007two}. An EM-like algorithm is often adopted to iteratively minimize the energy until convergence. Other works that are based on such hypothesis-and-test paradigm adopt spectral clustering in the final grouping step \cite{Chin2009,chin2010accelerated}. In contrast to the iterative approaches, analytic approaches are characterized by elegant mathematical formulation, the most well-known among them being those based on the sparsity \cite{Elhamifar2013} and low-rank \cite{liu2013robust} assumptions and their variants. Nonetheless, very few works \cite{Barath2018,goh2007segmenting,sugaya2004geometric,torr1998geometric} have considered the problem of fitting multiple model of various types, and in these few works, the types are assumed to be known a priori, well-defined, and cleanly separable, which is often not the case. 

\noindent\textbf{Deep Learning for Geometric Problems}:
Using deep learning to solve geometric model fitting has received growing considerations. The dense approaches start from raw image pairs to estimate models such as homography \cite{detone2016deep} or non-rigid transformation \cite{rocco2017convolutional}. \cite{melekhov2017relative} proposed to estimate the camera pose directly from image sequences. 
% Learning directly from image could be harmed by large baseline and occlusion \cite{yi2018learning}. %Moreover, they are all specifically developed for image data. A more general approach for geometric problems is still missing.

In contrast to the preceding works, DSAC\cite{brachmann2017dsac} learns to extract from sparse feature correspondences some geometric models in a manner akin to RANSAC. The ability to learn representations from sparse points was also developed recently\cite{Qi2017,QiLSG_NIPS2017}. This ability was exploited by \cite{yi2018learning} to fit camera motion (essential matrix) from noisy correspondences. Despite the promising results, none of the existing works have considered generic model fitting and, more importantly, fitting data of multiple models and even multiple types. In this work, we formulate the generic multi-model multi-type fitting problem as one of learning good representations for clustering.

\noindent\textbf{Deep Learning for Clustering}:
%Clustering was treated as an unsupervised learning problem with separate feature extraction and representation learning \cite{ding2004k}. With the advent of deep neural networks, simultaneous representation learning and clustering provides a better solution \cite{aljalbout2018clustering}. 
One line of researches tackle the problem by minimizing the reconstruction loss  \cite{huang2007unsupervised,vincent2008extracting,lee2009convolutional}.  The reconstruction loss can be further combined with various losses for achieving clustering objectives \cite{tian2014learning,yang2016joint,ji2017deep,xie2016unsupervised}. Among these, the k-means loss was proposed by \cite{yang2016joint} optimizing the points to cetner distance. \cite{xie2016unsupervised} proposed to minimize the KL-Divergence between the original feature and the embedded features. The locality-preserving loss \cite{huang2014deep} aims to find feature embedding conforming to a manifold constraint.%, and the group sparsity loss \cite{huang2014deep} was put forth to impose block diagonal structure of the similarity matrix.
 These works learn feature embedding in an unsupervised fashion. %It may succeed in clustering data with clear division (e.g. numbers and faces). However, 
When faced with the greater ambiguity envisaged in our multi-type fitting tasks, these unsupervised approaches are likely to face difficulties. %There are often more than one ways of clustering the data, and these clusters are not necessarily hierarchically arranged. 
For example, feature points on near planar foreground and background with large depth variation cannot be easily separated without proper assumption. %either as specifically designed new representation \cite{Li_2013_ICCV} or learning from labelled data. 

To take advantage of labelled data, metric learning has been applied to clustering \cite{Xing2003,hershey2016deep}. %Early study into distance metric for clustering proposed to learn a Mahalanobis distance for good clustering performance \cite{Xing2003}. 
With the advent of deep neural network, works in metric learning tend to focus on designing metric learning losses\cite{Chopra2005, Schroff2015, Sohn2016, hershey2016deep, song2017deep}. %Losses are proposed to learn good featur embeddings, including the constrastive loss \cite{Chopra2005}, triplet loss\cite{Schroff2015}, N-pair loss \cite{Sohn2016}. 
Among these, \cite{hershey2016deep} minimizes the L2 distance between the predicted and ground-truth affinities and provides a competitive baseline. To further take the global points distribution, we propose the clustering-specific loss MaxInterMinIntra, which optimize the inter-cluster separation and intra-cluster variance.

%\noindent\textbf{Contributions}: we summarize our contributions in this work as follows. 
%
%\begin{itemize}
%
%
%\item First, we formulate the traditional multi-model/class fitting problem in a supervised fashion. TO the best of our knowledge, this is the first attempt to use end-to-end learning approach for general multi-model fitting. 
%
%\item Second, we explored three? metric learning approaches to learning a new feature embedding where points belonging to different geometric models are easily separable. The metric learning approach also succeed to separate multiple instances of different underlying geometric models.
%
%\item Moreover, given explicit geometric model prior, we propose to incorporate an additional constraint, the fitting residual, to further regularize the learning process.
%
%\item Finally, we develop a simple Ransac with prior to refine the results obtained from the output of deep network.
%
%\end{itemize}

\section{Methodology}

In this section, we first explain the training process of our multi-model multi-type fitting network. We then introduce existing metric learning loss and our MaxInterMinIntra loss. %Then, for the testing phase, a RANSAC refinement step is presented to explicitly add geometric constraints if desired.

\begin{figure}[!ht]
\begin{center}
\includegraphics[width=1\linewidth]{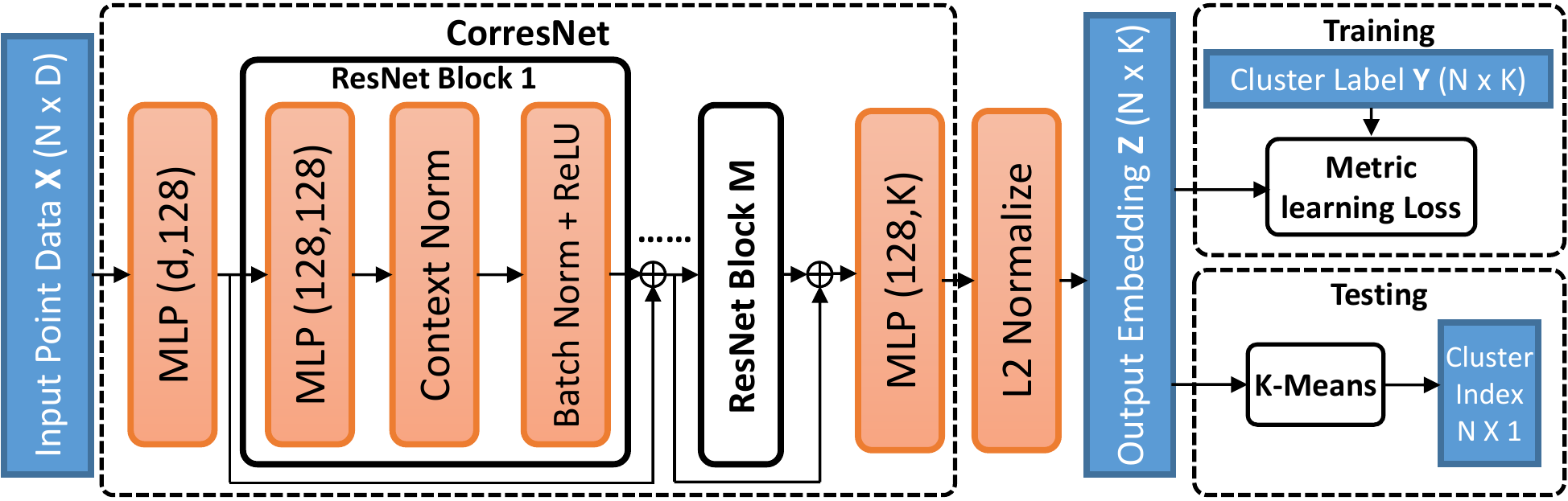}
\caption{Our multi-model multi-type fitting network. We adopt the same cascaded CorresNet blocks as \cite{yi2018learning}. The metric learning loss is defined to learn good feature representation. }\label{fig:Network}
\end{center}\vspace{-0.5cm}
\end{figure}

\subsection{ Network Architecture}

We denote the input sparse data with $N$ points as $\matr{X}=\{\vect{x}_{i}\}_{i=1\cdots N}\in\set{R}^{D\times N}$ where each individual point is $\vect{x}_{i}\in\set{R}^D$. The input sparse data could be geometric shapes, feature correspondences in two frames or feature trajectories in multiple frames. We further denote the one-hot key encoded labels accompanying the input data as $\matr{Y}=\{\vect{y}_{i}\}\in\{0,1\}^{K\times N}$ where $\matr{y}_{i}\in \{0,1\}^{K}$ and $K$ is the number of clusters or partitions of the input data. 

Cascaded multi-layer perceptrons (mlps) has been used to learn feature representation from generic point input \cite{Qi2017,yi2018learning}. We adopt a backbone network similar to CorresNet \cite{yi2018learning}\footnote{Alternative sparse data networks, e.g. PointNet \cite{Qi2017}, are applicable as well} shown in Fig.~\ref{fig:Network} . The output embedding of the CorresNet is denoted as $\matr{Z}=\{f(\matr{X};\Theta)\}\in\set{R}^{K\times N}$. To make the output $\matr{Z}$ clustering-friendly, we apply a differentiable, clustering-specific loss function $\mathcal{L}(\matr{Z},\matr{Y})$,  
measuring the match of the output feature representation with the ground-truth labels.  The problem now becomes that of learning a CorresNet backbone $f(\matr{X};\Theta)$ that minimizes the loss $\mathcal{L}(\matr{Z},\vect{Y};\Theta)$.

%Finally, the objective is to learn the model 
%
%We empirically discover that the repetition of ResNet block yields best performance at 30-60 times. 

\subsection{Clustering Loss}
We expect our clustering loss function to have the following characteristics. First, It should be invariant to permutation of models, e.g. the order of these models are exchangeable. Second the loss must be adaptable to varying number of groups. Lastly, the loss should enable good separation of data points into clusters. We consider the following loss functions. 

\noindent\textbf{L2Regression Loss}:
Given the ground-truth labels $\vect{Y}$ and the output embeddings $\matr{Z}=f(\vect{X};\Theta)$, the ideal and reconstructed affinity matrices are respectively,
\begin{equation}
\begin{split}
&\matr{K} = \vect{Y}^\top\vect{Y}, \quad \matr{\hat{K}} = \matr{Z}^\top\matr{Z}
\end{split}
\end{equation}

The training objective is to minimize the difference between $\matr{K}$ and $\matr{\hat{K}}$ measured by element-wise L2 distance \cite{hershey2016deep}.
\vspace{-0.2cm}
\begin{equation}
\begin{split}
L(\Theta) &= ||\matr{K}-\matr{\hat{K}}||_F^2 \\
&= ||\matr{Y}^\top\matr{Y} - \vect{Z}^\top\vect{Z}||_F^2\\
&= ||f(\matr{X};\Theta)^\top f(\matr{X};\Theta)||_F^2 - 2||f(\matr{X};\Theta)\vect{Y}^\top||_F^2
\end{split}
\end{equation}

The above L2 Regression loss is obviously differentiable w.r.t. $f(\matr{X};\Theta)$. Since the output embedding $\matr{Z}$ is L2 normalized, the inner product between two point representations is $\vect{z}_i^\top\vect{z}_j\in[-1,1]$. 
% The L2 distance gives the same penalty to $\vect{z}_i^\top\vect{z}_j=0.1$ and $\vect{z}_i^\top\vect{z}_j=-0.1$ with ground-truth affinity % $\vect{y}_i^\top\vect{y}_j=0$. Whereas, the latter obviously gives a larger margin between $\vect{z}_i$ and $\vect{z}_j$. 
% To further encourage separating points of distinct groups, we introduce a variant of the L2 Regression loss in the following.

\noindent\textbf{Cross-Entropy Loss}: As alternative to the L2 distance, one could measure the discrepancy between $\matr{K}$ and $\hat{\matr{K}}$ as KL-Divergence. Since $D_{kl}(\matr{K}||S(\matr{\hat{K}}))=H(\matr{K},S(\matr{\hat{K}})) - H(\matr{K})$, where $H(\cdot)$ is the entropy function and $S(\cdot)$ is the sigmoid function, with fixed $\matr{K}$, we simply need to minimize the cross-entropy $H(\matr{K},S(\matr{\hat{K}}))$ which derives the following element-wise cross-entropy loss,
\vspace{-0.2cm}

\begin{equation}
\begin{split}
L(\Theta)&= \sum_{i,j} H\left(\vect{y}_{i}^\top\vect{y}_{j},S\left(\vect{z}_{i}^\top\vect{z}_{j}\right)\right) \\
&= \sum_{i,j} H(\vect{y}_{i}^\top\vect{y}_{j},S(f(\vect{x}_i;\Theta)^\top f(\vect{x}_i;\Theta)))
\end{split}
\end{equation}

\vspace{-0.2cm}
 The cross-entropy loss is more likely to push points $i$ and $j$ of the same cluster together faster than L2Regression, i.e. inner product $\vect{z}_i^\top\vect{z}_j \rightarrow 1$ and those of different clusters apart, i.e. inner product $\vect{z}_i^\top\vect{z}_j \rightarrow -1$.

\noindent\textbf{MaxInterMinIntra Loss}:
Both the above losses consider the pairwise relation between points; the overall point distribution in the output embedding is not explicitly considered. We now propose a new loss which takes a more global view of the point distribution rather than just the pairwise relations. Specifically, we are inspired by the classical Fisher LDA \cite{fisher1936use}. LDA discovers a linear mapping ${z}=\vect{w}^\top\vect{x}$ that maximizes the distance between class centers/means $\mu_i=1/N \sum_j {z}_j$ and minimizes the scatter/variance within each class $s_i = \sum_j(z_j-\mu_i)^2$. Formally, the objective for a two-class problem is written as,
\begin{equation}
J(\vect{w})=  \frac{|\mu_1-\mu_2|^2}{s_1^2+s_2^2}
\end{equation}
which is to be maximized over $\vect{w}$. 
For linearly non-separable problem, one has to design kernel function to map the input features before applying the LDA objective. Equipped now with more powerful nonlinear mapping networks, we adapt the LDA objective---for the multi-class scenarios---to perform these mappings automatically as below, 
\begin{equation}
J(\Theta)=\frac{\min\limits_{m,n\in\{1\cdots K\}, m\neq n}||\bm{\mu}_m - \bm{\mu}_n||^2_2}{\max\limits_{l\in\{1\cdots K\}} s_l}
\end{equation}
where $\bm{\mu}_m=\frac{1}{|\mathcal{C}_m|}\sum_{i\in\mathcal{C}_m} \vect{z}_i$, $s_l = \sum_{i\in\mathcal{C}_l}||\vect{z}_i-\bm{\mu}_l||^2_2$ and $\mathcal{C}_l$ indicating the set of points belonging to cluster $l$.  We use the extremas of the inter-cluster distances and intra-cluster scatters (see Fig.~\ref{fig:MaxInterMinIntra}) so that the worst case is explicitly optimized. Hence, we term the loss as MaxInterMinIntra (MIMI). By applying log operation on the objective,% and weighing the relative importance between the two terms, 
we arrive at the following loss function to be minimized: 

\begin{figure}
\includegraphics[width=1\linewidth]{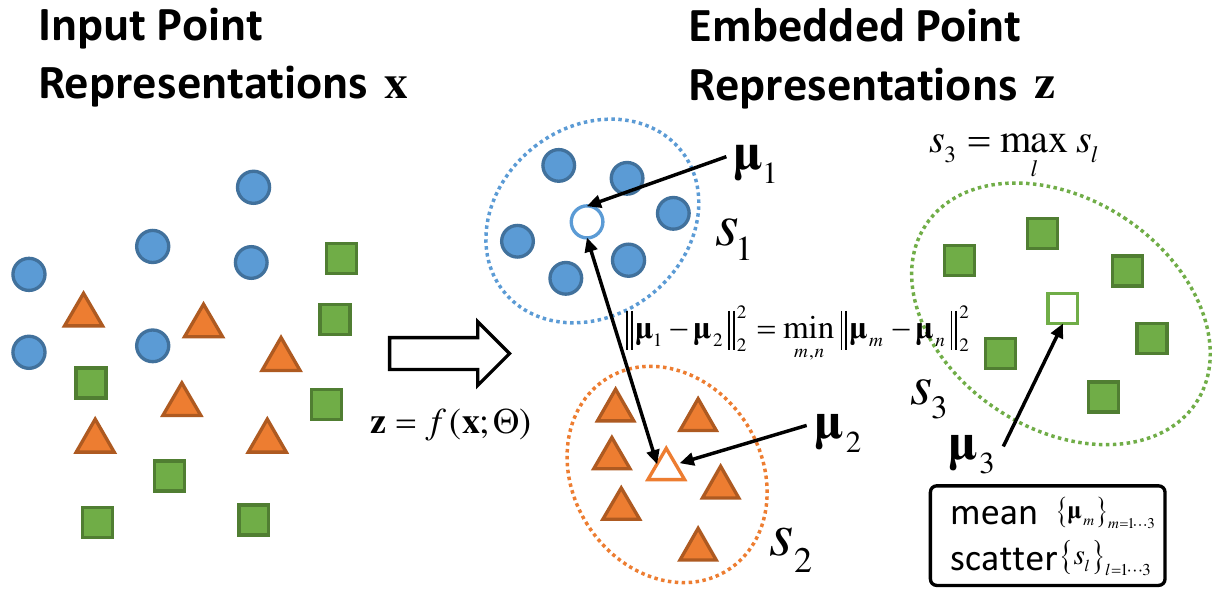}
\caption{Illustration of MaxInterMinIntra loss for point representation metric learning. The objective considers the minimal distance $\min_{m,n}||\bm{\mu}_m-\bm{\mu}_n||_2^2$ between clusters and maximal scatter  $\max_l s_l$ within clusters.}\label{fig:MaxInterMinIntra}
\vspace{-0.5cm}
\end{figure}

\begin{equation}\label{eq:MaxInterMinIntra}
L(\Theta) = -\log \min\limits_{m,n} ||\bm{\mu}_m-\bm{\mu}_n||_2^2 + \log\max_{l} s_l
\end{equation}
\noindent
One can easily verify that the MaxInterMinIntra loss is differentiable w.r.t. $\vect{z}_i$. We give the gradient in Eq~(7). %with $\hat{m}$,$\hat{n}$ and $\hat{l}$ indicating the corresponding indices of clusters.???

\noindent\textbf{Optimization}:
The Adam optimizer \cite{kingma2014adam} is used to minimize the loss $L(\Theta)$. The learning rate is fixed at $1e-4$ and mini-batch at one frame pair or sequence. The mini-batch size cannot exceed one because the number of points/correspondences is not uniform across different sequences. For all tasks, we train the network 300 epochs. %For both L2 Regression and CrossEnt loss, there are no hyper-parameters, 

%\begin{equation}
%\resizebox{0.98\hsize}{!}
%{$
%\begin{split}
%&\nabla_\Theta l(\Theta) =\\
%& -\sum\limits_{i\in\mathcal{C}_m}\frac{\frac{1}{|\mathcal{C}_m|^2}\left(2\vect{z}_i + \sum\limits_{j\in\mathcal{C}_m,j\neq i} \vect{z}_j\right) - \frac{1}{|\mathcal{C}_m||\mathcal{C}_n|}\sum\limits_{j\in\mathcal{C}_n}\vect{z}_j}{||\bm{\mu}_m-\bm{\mu}_n||_2^2}\nabla_\Theta f(\matr{X}_i;\Theta)\\
%& -\sum\limits_{j\in\mathcal{C}_n}\frac{\frac{1}{|\mathcal{C}_n|^2}\left(2\vect{z}_j + \sum\limits_{i\in\mathcal{C}_m,i\neq j} \vect{z}_i\right) - \frac{1}{|\mathcal{C}_n||\mathcal{C}_m|}\sum\limits_{k\in\mathcal{C}_m}\vect{z}_k}{||\bm{\mu}_m-\bm{\mu}_n||_2^2}\nabla_\Theta f(\matr{X}_j;\Theta)\\
%&+\alpha\sum\limits_{k\in \mathcal{C}_l} \Big(2\vect{z}_k - \frac{1}{|\mathcal{C}_l|}\left(2\vect{z}_k + \sum\limits_{j\in\mathcal{C}_l,j\neq i}\vect{z}_j\right) +\\
%& \frac{1}{|\mathcal{C}_l|^2}\left(2\vect{z}_k + 2\sum\limits_{j\in\mathcal{C}_l,j\neq i}\vect{z}_j\right)\Big)\nabla_\Theta f(\matr{X}_k;\Theta)
%\end{split}
%$}
%\end{equation}

\begin{figure*}[!htb]
\begin{equation}
\resizebox{0.92\hsize}{!}
{$
\begin{split}
&\nabla_\Theta L(\Theta) = -\sum\limits_{i\in\mathcal{C}_m}\frac{\frac{1}{|\mathcal{C}_m|^2}\left(2\vect{z}_i + \sum\limits_{j\in\mathcal{C}_m,j\neq i} \vect{z}_j\right) - \frac{1}{|\mathcal{C}_m||\mathcal{C}_n|}\sum\limits_{j\in\mathcal{C}_n}\vect{z}_j}{||\bm{\mu}_m-\bm{\mu}_n||_2^2}\nabla_\Theta f(\matr{x}_i;\Theta) -\sum\limits_{j\in\mathcal{C}_n}\frac{\frac{1}{|\mathcal{C}_n|^2}\left(2\vect{z}_j + \sum\limits_{i\in\mathcal{C}_m,i\neq j} \vect{z}_i\right) - \frac{1}{|\mathcal{C}_n||\mathcal{C}_m|}\sum\limits_{k\in\mathcal{C}_m}\vect{z}_k}{||\bm{\mu}_m-\bm{\mu}_n||_2^2}\nabla_\Theta f(\matr{x}_j;\Theta)\\
&+\alpha\sum\limits_{k\in \mathcal{C}_l} \Big(2\vect{z}_k - \frac{1}{|\mathcal{C}_l|}\left(2\vect{z}_k + \sum\limits_{j\in\mathcal{C}_l,j\neq i}\vect{z}_j\right) + \frac{1}{|\mathcal{C}_l|^2}\left(2\vect{z}_k + 2\sum\limits_{j\in\mathcal{C}_l,j\neq i}\vect{z}_j\right)\Big)\nabla_\Theta f(\matr{x}_k;\Theta)
\end{split}
$}
\end{equation}\vspace{-0.5cm}
\end{figure*}%\label{eq:MIMI_Gradient}

\subsection{Inference}
%The above network takes the label of each cluster/group of points as input and learns a good point feature embedding for separation. 
% As with the conventional approaches towards multi-model fitting \cite{Magri2014,Isack2012}, we predict cluster labels for testing data $\{\vect{x}_j\}_{j=1\cdots N_{te}}$, i.e. the data to be fitted. Points belong to the same model, e.g. line, circle, plane or rigid motion, are expected to be assigned the same cluster label. To achieve this
During testing, we apply standard K-means to the output embeddings $\{\vect{z}_j\}_{j=1\cdots N_{te}}$. 
% Thanks to the objective of embedding points into tight cluster with low variance, K-means has a high chance of finding a correct clustering of testing data. 
This step is applicable to both multi-model and multi-type fitting problems, as we do not need to specify explicitly the type of model to fit. Finally, with unknown number of models $K$, we propose to analyze the K-means residuals, 
\begin{equation}
r(K) = \sum_{m=1\cdots K}\sum_{i\in\mathcal{C}_m} ||\vect{z}_i - \bm{\mu}_m||_2^2
\end{equation}

Good estimate of $K$ often yields low $r(K)$ and further increasing $K$ does not significantly reduce $r(K)$. So we find the $K$ at the `elbow' position. We adopt two off-the-shell approaches for this purpose, the second order difference (SOD)\cite{Zhang2012} and silhouette analysis \cite{rousseeuw1987silhouettes}. Both are parameter-free.

\section{Experiment}

We demonstrate the performance of our network on both synthetic and real world data, with extensive comparisons with traditional geometric model fitting algorithms. Our focus is on the multi-type setting (the first two experiments on LCE and KT3DMoSeg), but we also carry out experiments on the pure multi-model scenario (LCE-unmixed and Adelaide RMF) experiments. 

\subsection{Datasets}

% We evaluate our multi-model multi-type fitting network on both synthetic data and real world tasks. Both the multi-modek single-type and multi-type are separately evaluated for extensive comparison. In specific, we have two groups of experiments. For the first group, we evaluate the multi-type fitting on two datasets as below,

\noindent\textbf{Synthesized Lines, Circles and Ellipses (LCE)}: Fitting ellipses has been a fundamental problem in computer vision \cite{fitzgibbon1999direct}. We synthesize for each sample four different types of conic curves in a 2D space, specifically, one straight line, two ellipses and one circle. We randomly generate 8,000 training samples, 200 validation samples and 200 testing samples. Each point is perturbed by adding a gaussian noise with $\sigma=0.05$.

\noindent\textbf{KT3DMoSeg} \cite{xu2018motion}: This benchmark was created based upon the KITTI self-driving dataset\cite{Geiger2013IJRR} with 22 sequences in total. %Dense points are tracked across multiple frames in each sequence and labelled according to each object and background. 
Each sequence contains two to five rigid motions. As analyzed by \cite{xu2018motion}, the geometric model for each individual motion can range from an affine transformation, a homography, to a fundamental matrix, with no clear dividing line between them. We evaluate this benchmark to demonstrate our network's ability to tackle multi-model multi-type fitting. For fair comparison with all existing approaches, we only crop the first 5 frames of each sequence for evaluation, so that the broken trajectory does not give undue advantage to certain methods.

% For the next group, we demonstrate the ability on single-type multi-model fitting tasks with two datasets as below.

\noindent\textbf{Synthesized Lines, Circles and Ellipses Unmixed (LCE-Unmixed)}: To demonstrate the ability of our network on single-type multi-model fitting, we also randomly generate in each sample a single class of conic curves in 2D space (lines, circles, or ellipses) but with multiple instances (2-4) of them. The number of training, validation and testing samples are the same as those of the multi-type LCE setting. Same perturbation as LCE is applied here.

\noindent\textbf{Adelaide RMF Dataset} \cite{wong2011dynamic}: This dataset consists of 38 frame pairs, of which half are designed for multi-model fitting (the model being homographies induced by planes). The number of planes is between two to seven. The other 19 frame pairs are designed for two-view motion segmentation. It is nominally a single-type multiple fundamental matrix fitting problem and has been treated as such by the community. While we put the results under the single-type category, we hasten to add that there might indeed be degeneracies, i.e. near planar rigid objects, (and hence mixed types) present in this dataset, no matter how minor. The number of motions is between one to five.

\subsection{Multi-Type Curve Fitting}

The multiple types in this curve fitting task comprises of lines, circles, and ellipses in the LCE dataset. Note that there is no clear dividing boundary between them as they can be all explained by the general conic equation (with the special cases of lines and circles obtained by setting some coefficients to $0$):

\begin{equation}\label{eq:Conic}
Ax^2+Bxy+Cy^2+Dx+Ey+F=0
\end{equation}

There are two ways to adapt the traditional multi-model methods for this multi-type setting. One approach is to formulate the multi-type fitting problem as fitting multiple models parameterized by the same conic equation in Eq~(\ref{eq:Conic}). This approach is termed \textit{HighOrder} (H.O.) fitting.
Alternatively, one could sequentially fit three types of models, which is termed \textit{Sequential} (Seq.) fitting. For ellipse-specific fitting, the direct least square approach \cite{fitzgibbon1999direct} is adopted.
For our model, we evaluate the various metric learning losses introduced in Section~3.2 and present the results in Tab.~\ref{tab:SyntheticMultiType}. 
The results are reported with the optimal setting determined by the validation set. We evaluate the performance by two clustering metrics, Classification Error Rate (Error Rate), i.e. the best classification results subject to permutation of clustering labels, and Normalize Mutual Information (NMI). Comparisons are made with state-of-the-art multi-model fitting algorithms including T-linkage \cite{Magri2014}, RPA \cite{Magri2015} and RansaCov \cite{magri2016multiple}. We notice that T-linkage returns extremely over-segmented results in the sequential setting, e.g. more than 10 lines, making classification error evaluation intractable as it involves finding the permutation label with lowest error rate. For our model, we evaluate the three loss variants, the L2 Regression loss (L2), Cross Entropy loss (CE) and MaxInterMinIntra loss (MIMI).

%% Table generated by Excel2LaTeX from sheet 'Export'
%\begin{table}[htbp]
%  \centering
%\caption{Evaluations on synthetic multi-model and multi-type fitting dateset. $\uparrow$ and $\downarrow$ indicate the number is the higher or lower the better respectively. $-$ indicates evaluation intractable.}
%  \setlength\tabcolsep{2.8pt} % default value: 6pt
%  \resizebox{0.5\linewidth}{!}{
%    \begin{tabular}{cp{6em}ll}
%    \toprule
%    \multicolumn{2}{p{5.43em}}{\textbf{Model}} & \multicolumn{1}{p{2.93em}}{\textbf{Error}$\downarrow$} & \multicolumn{1}{p{3.145em}}{\textbf{NMI}$\uparrow$} \\
%    \midrule
%    \multirow{2}[2]{*}{\begin{sideways}H.O.\end{sideways}} & T-Linkage\cite{Magri2014} &      52.14 & 0.3404 \\
%          & RPA\cite{Magri2015}   & 39.43 & 0.4641 \\
%\cmidrule{2-4}    \multirow{2}[2]{*}{\begin{sideways}Seq.\end{sideways}} & T-Linkage\cite{Magri2014} &     - & - \\
%          & RPA\cite{Magri2015}   & 23.17 & 0.6574 \\
%    \midrule
%    \multirow{4}[4]{*}{\begin{sideways}Our Models\end{sideways}} & \multicolumn{1}{l}{\textbf{Loss}} &       &  \\
%\cmidrule{2-4}          & L2Regress & 18.49 & 0.7126 \\
%          & CrossEnt & 18.32 & 0.7103 \\
%          & MIMI  & \textbf{18.04} & \textbf{0.7272} \\
%    \bottomrule
%    \end{tabular}%
%    }
%  \label{tab:SyntheticMultiType}%
%\end{table}%

% Table generated by Excel2LaTeX from sheet 'Export'
\begin{table}[htbp]
  \centering
\caption{Evaluations on synthetic multi-model and multi-type fitting dataset. $\uparrow$ and $\downarrow$ indicate the number is the higher or lower the better respectively. $-$ indicates evaluation intractable.}
   \setlength\tabcolsep{4pt} % default value: 6pt
  \resizebox{1\linewidth}{!}{
    \begin{tabular}{p{2.3em}ccccccccc}
    \toprule
    \multirow{2}[4]{*}{\textbf{Mdl.}} & \multicolumn{2}{c}{T-Linkage\cite{Magri2014}} & \multicolumn{2}{c}{RPA\cite{Magri2015}} & \multicolumn{2}{c}{RansaCov\cite{magri2016multiple}} & \multicolumn{3}{c}{Our Models} \\
\cmidrule(lr){2-3}\cmidrule(lr){4-5}\cmidrule(lr){6-7} \cmidrule{8-10}   \multicolumn{1}{c}{} & \multicolumn{1}{c}{H.O.} & \multicolumn{1}{c}{Seq.} & \multicolumn{1}{c}{H.O.} & \multicolumn{1}{c}{Seq.} & \multicolumn{1}{c}{H.O.} & \multicolumn{1}{c}{Seq.} & \multicolumn{1}{c}{L2} & \multicolumn{1}{c}{CE} & \multicolumn{1}{c}{MIMI} \\
    \midrule
    \textbf{Err}$\downarrow$ & 52.14 & -  & 39.43 & 23.17 & 40.57 & 24.04 & 18.49 & 18.32 & \textbf{18.04} \\
    \textbf{NMI}$\uparrow$ & 0.340 & - & 0.464 & 0.667 & 0.394 & 0.604 & 0.713 & 0.720 & \textbf{0.727} \\
    \bottomrule
    \end{tabular}%
    }
  \label{tab:SyntheticMultiType}%
\end{table}%

We make the following observations about the results. First, all our metric learning variants outperform the \textit{HighOrder} and \textit{Sequential} multi-type fitting approaches. Second, the all-encompassing model used in the \textit{HighOrder} approach suffers from ill-conditioning when fitting simpler models. Thus, the performance is much inferior to that of \textit{Sequential} fitting. However, it is worth noting that  despite the \textit{Sequential} approach being given the strong a priori knowledge of both the model type and the number of model for each type, its performance is still significantly worse off than ours. 

% PEaRL \cite{Isack2012}. For our model, we evaluate first vanilla metric learning models including all alternative metric learning losses introduced in Section~3.2. %Then we report the results with geometric prior refinement introduced in Section~3.3. The comparisons are presented in Tab.~\ref{tab:MultiLineMultiClass}.

\begin{figure}
\includegraphics[width=1\linewidth]{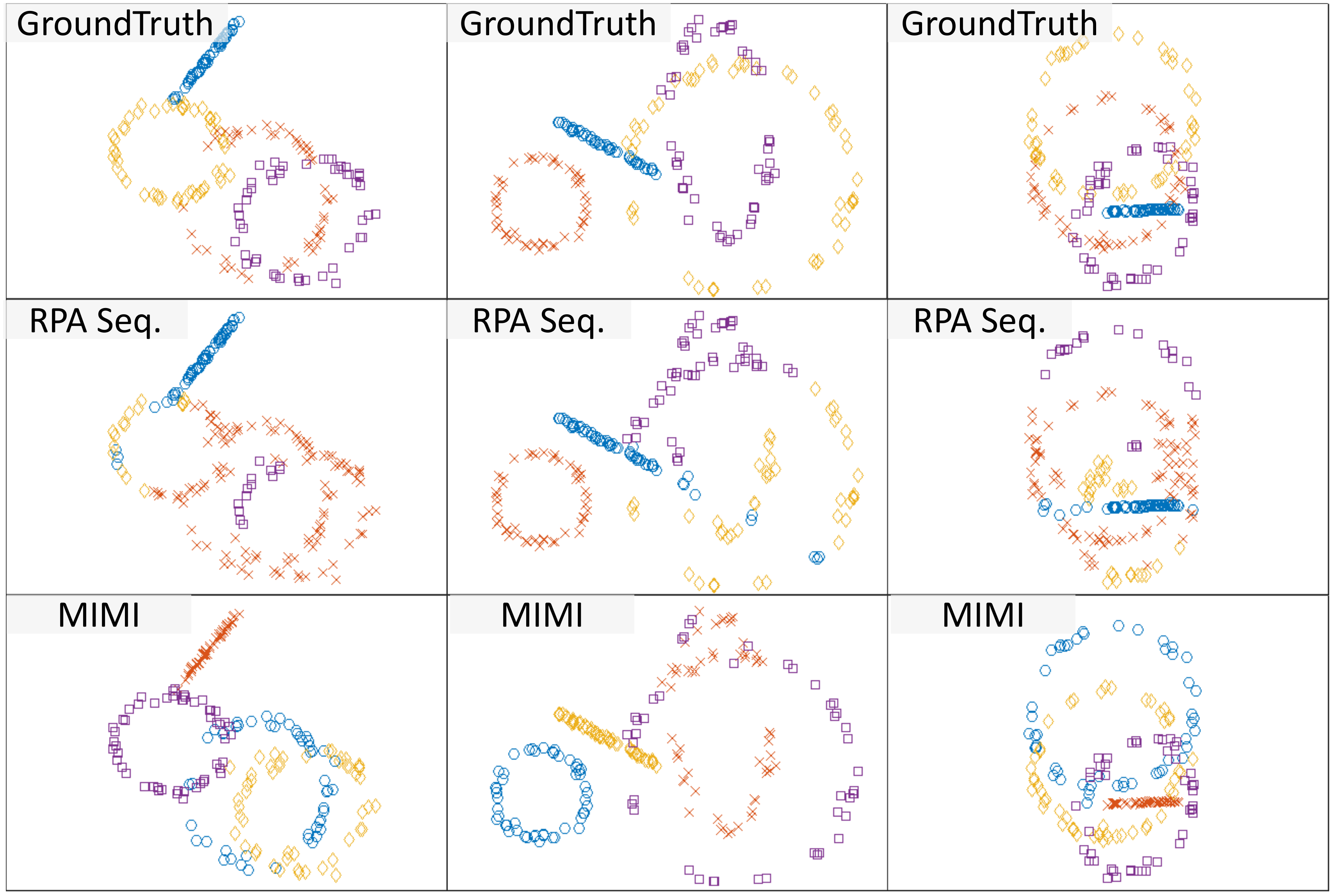}
\caption{Examples of multi-type fitting on synthetic dataset. We only show the RPA results based on the \textit{Sequential} fitting approach.}\label{fig:MultiTypeExamples}
\vspace{-0.3cm}
\end{figure}

For qualitative comparison, we visualize the ground-truth and segmentation results of each method in Fig.~\ref{fig:MultiTypeExamples}. Our clustering results on the bottom row show success in discovering all individual shapes with mistakes made only at the intersections of individual structures. Though good at separating straight line, the RPA failed to discover ellipses as sampling all 5 inliers amidst the large number of outliers and fitting an ellipse from even correct 5 support points with noise (noise in coordinate) are both very difficult, the latter demonstrated in \cite{fitzgibbon1999direct}. %(You didn’t explain the noise generation process in the LCE description).

\subsection{Multi-Type Motion Segmentation}

\begin{figure*}[!htb]
\centering
\includegraphics[width=1\linewidth]{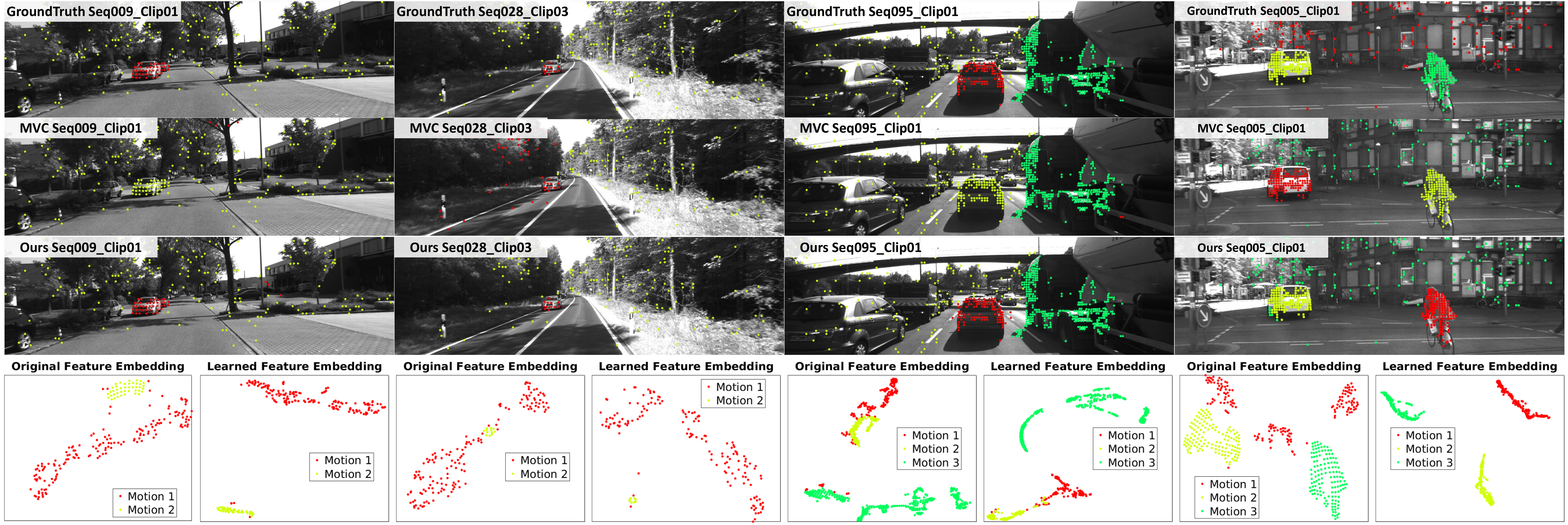}
\caption{Qualitative comparison on 4 sequences from KT3DMoSeg. First row are the ground-truth. Second and third rows are the results of Multi-View Clustering \cite{xu2018motion} and our multi-type network respectively. The last row are the point feature embeddings before and after learning.}\label{fig:KT3DMoSeg}\vspace{-0.4cm}
\end{figure*}

The KT3DMoSeg benchmark \cite{xu2018motion} is put forth for the task of motion segmentation. Each sequence often consists of a background whose motion can be  explained in general by a fundamental matrix while the models for the foreground motions can sometimes be ambiguous due to the limited spatial extent of the objects, thus giving rise to mixed types of models. For example, in Fig.~\ref{fig:KT3DMoSeg}, the vehicles in `Seq009\_Clip01' and `Seq028\_Clip03' can be roughly explained by an Affine transformation or Homography while the oil tanker in `Seq095\_Cip01' should be modeled by a fundamental matrix. Even the background motion can be ambiguous to model, when the background is dominated by a plane, for instance, the quasi-planar row of trees on the right side of the road in `Seq028\_Clip03' is likely to lead to degeneracies in the fundamental matrix estimation and thus cause errors in the traditional method (second row). For this dataset, we use the first five frames of each sequence for fair comparison and apply leave-one-out cross-validation, i.e. repeatedly train on 21 sequences and test on the left-out sequence; we dubbed this  the `Vanilla' setting. Each sequence has between 10-20 frames, so we could further increase the training data by augmenting with all the remaining five-frame clips from each sequence with no overlap; this is termed as the `Augment' setting. The testing clips (first five frames of each sequence) are kept the same for both settings. We compare with subspace clusering approaches, GPCA\cite{vidal2005generalized}, LSA\cite{Yan2006}, ALC\cite{Rao2010}, LRR\cite{liu2013robust}, MSMC\cite{Dragon2012} and SSC\cite{Elhamifar2013} and the multi-view clustering (MVC) methods in \cite{xu2018motion}. Results are presented in Tab.~\ref{tab:KT3DMoSeg}.

% Table generated by Excel2LaTeX from sheet 'KT3DMoSeg'
\begin{table*}[!htbp]
  \centering
   \caption{Motion segmentation performance on KT3DMoSeg 5-frame task. Three losses, L2 Regress (L2), Cross Entropy (CE) and MaxInterMinIntra (MIMI) are evaluated. Numbers are in $\%$.}
     \footnotesize
   \setlength\tabcolsep{2pt} % default value: 6pt
	  \resizebox{0.75\linewidth}{!}{
    \begin{tabular}{p{5em}ccccccccccccc}
    \toprule
    \multicolumn{1}{c}{\multirow{3}[6]{*}{\textbf{Model}}} & \multicolumn{7}{c}{\textbf{State-of-the-Arts}}                 & \multicolumn{6}{c}{\textbf{Our Models}} \\
\cmidrule(lr){2-8}\cmidrule(lr){9-14}    \multicolumn{1}{c}{} & \multicolumn{1}{c}{\multirow{2}[4]{*}{GPCA\cite{vidal2005generalized}}} & \multicolumn{1}{c}{\multirow{2}[4]{*}{ALC\cite{Rao2010}}} & \multicolumn{1}{c}{\multirow{2}[4]{*}{LSA\cite{Yan2006}}} & \multicolumn{1}{c}{\multirow{2}[4]{*}{LRR\cite{liu2013robust}}} & \multicolumn{1}{c}{\multirow{2}[4]{*}{MSMC\cite{Dragon2012}}} & \multicolumn{1}{c}{\multirow{2}[4]{*}{SSC\cite{Elhamifar2013}}} & \multicolumn{1}{c}{\multirow{2}[4]{*}{MVC\cite{xu2018motion}}} & \multicolumn{3}{c}{\textbf{Vanila}} & \multicolumn{3}{c}{\textbf{Augment}} \\
\cmidrule(lr){9-11}\cmidrule(lr){12-14}    \multicolumn{1}{c}{} &       &       &       &       &       &       &       & L2    & CE    & MIMI  & L2    & CE    & MIMI \\
    \midrule
    Mean Err & 36.46 & 15.17 & 36.3  & 22.00 & 32.74 & 26.62 & \textbf{10.99} & 14.04    & 12.44 & 14.15 & 8.87  & 11.48 & \textbf{6.78} \\
    Med. Err & 33.93 & 16.42 & 40.3  & 18.16 & 36.48 & 29.14 & \textbf{6.57}  & 8.90  & 9.87  & 10.89 & {6.68} & 9.42  & \textbf{2.67} \\
    \bottomrule
    \end{tabular}%
  \label{tab:KT3DMoSeg}%
  }
\end{table*}%

We make the following observations about the results. Our vanilla leave-one-out approach achieved very competitive performance on all 22 sequences in KT3DMoSeg. In the `Augment' setting, our approach even outperforms the state-of-the-art multi-view clustering approaches (MVC) \cite{xu2018motion}. Of all benchmark methods, only MVC has considered the multi-type fitting issue. However, the multi-view fusion proposed therein still does not guarantee that each rigid motion is explained by the correct model. Furthermore, we notice that our proposed MIMI metric is comparable to both the L2 Regression and cross entropy loss and gives even lower error when augmented with additional data. This suggests that optimizing the distribution of the embedded features with a clustering-specific loss is effective.

Finally, we present qualitative comparison between the results of MVC and ours in Fig.~\ref{fig:KT3DMoSeg}. Not only is the proposed network capable of correctly segmenting the aforementioned degenerate motions, it surpasses our expectations in how it performs in `Seq009\_Clip01'. Here the independently moving car (the yellow group in the ground truth image) has a flow field that is consistent with the epipolar constraint associated with the background motion (due to them both translating in the same direction) \cite{xu2018motion}. Without resorting to reconstructing the depth of the car, it would be impossible to separate it from the background. However, criteria involving depth would be very unwieldy to specify analytically in the existing approaches. Here, without having any preconceived notion of the geometrical model, our network seems to have learnt the requisite criteria to separate the independent motion. 
% The four sequences are difficult for geometric approaches due to the foreground motion staying close the translation along the epipolar line \cite{xu2018motion} and the object is too small to be captured by limited RANSAC trials. Explicit modeling with fundamental matrix allow them to be undetected. However, with the learning approaches, this issue is solved as the segmentation is learned from other sequences with similar motions. %The potential of multi-type network could further boosted with more available training data. 

\subsection{Multi-Model Fitting}

%As discussed, our multi-type network outperforms state-of-the-art Ransac-based and subspace clustering approaches in multi-type fitting tasks. 
In this section, we further demonstrate the ability of our network to handle conventional (i.e., single-type) multi-model fitting problems.

\begin{figure}[!htb]
\centering
\includegraphics[width=0.85\linewidth]{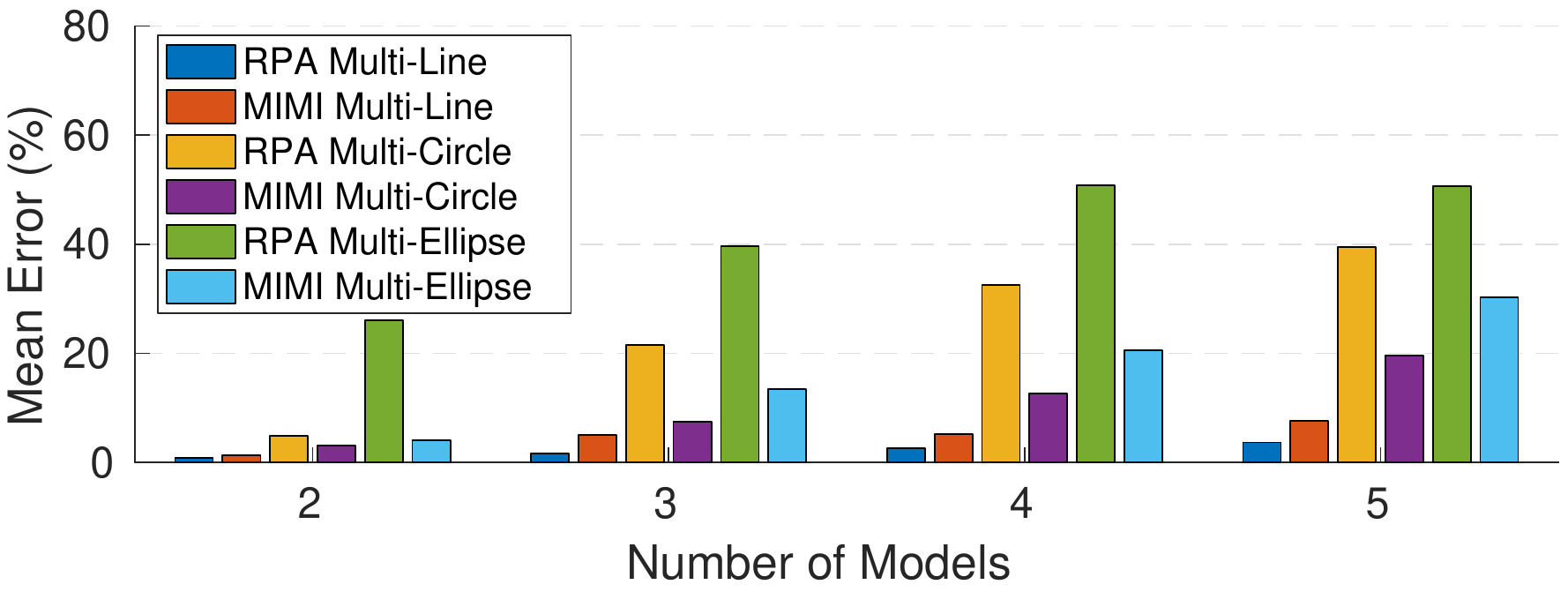}
\vspace{-0.2cm}
\caption{Performance v.s. the number of models for synthetic multi-model fitting.}\label{fig:SyntheticMultiModel}
\vspace{-0.2cm}
\end{figure}

\noindent\textbf{Synthetic Multi-Model Fitting}: In this experiment, we evaluate multi-model fitting of a single type (the type being line, circle or ellipse).  We adopt a similar training and testing split as in the synthetic LCE task, i.e. 8,000 training samples and 200 testing samples and compare with RPA\cite{Magri2015}. The results are presented in Fig.~\ref{fig:SyntheticMultiModel}. We conclude from the figure that, first, our multi-model network  performs comparably with RPA on multi-line segmentation task while outperforming RPA with large margin on the more challenging multi-circle and multi-ellipse segmentation tasks. Moreover, the performance drops sharply (higher error) from multi-line (blue) to multi-ellipse (green) fitting for RPA, with the drop getting more acute as the number of model increases. This suggests that the increasing size of the minimal support set (2 points for line, 3 points for circle and 5 points for ellipse) introduces great challenge for the Ransac-based approaches due to sampling imbalance. Hitting the true model becomes very difficult for model with larger support set and experiencing higher noise level. It is evident that our multi-model network is less sensitive to the complexity of the model, as the drop in performance (purple and cyan bars) are less significant.  Fig.~\ref{fig:SyntheticMultiModel} thus demonstrates that our deep learning approach is better able to deal with sampling imbalance, probably by picking up and leveraging on the additional regularity in the way the points are distributed.

\vspace{-0.2cm}
 %The performance of our multi-type network is still able to outperform state-of-the-art multi-model fitting approaches, however with a smaller margin. This is due to the known model type makes sampling and testing more robust.

%\begin{center}
%\begin{figure}
%\subfloat[Multi-Line]{\includegraphics[width=0.33\linewidth]{./Figure/MultiLinesPerfVsNumModel.pdf}}
%\subfloat[Multi-Circle]{\includegraphics[width=0.33\linewidth]{./Figure/MultiCirclesPerfVsNumModel.pdf}}
%\subfloat[Multi-Ellipse]{\includegraphics[width=0.33\linewidth]{./Figure/MultiEllipsesPerfVsNumModel.pdf}}
%\caption{Performance v.s. the number of models.}
%\end{figure}
%\end{center}

\noindent\textbf{Two-View Multi-Model Fitting}:
Finally, we evaluate the multi-model fitting task on the Adelaide RMF dataset \cite{wong2011dynamic}. For both the multi-planar and motion segmentation tasks, we carry out a leave-one-out cross-validation. For fair comparison, we report the classification error rate (ErrorRate). The state-of-the-art models being compared include J-Linkage \cite{toldo2008robust}, T-Linkage\cite{Magri2014}, RPA \cite{Magri2015}, RCMSA \cite{pham2014random} and ILP-RansaCov\cite{magri2016multiple}. The comparisons are presented in Tab.~\ref{tab:AdelaideRMF}. We observe that our multi-model network gives very competitive results on both the multi-planar and motion segmentation tasks. For the former task, our proposed MaxInterMinIntra (MIMI) loss yields $17.33\%$ which is better than many benchmark models. For the motion segmentation task, our model  with L2 Regression loss gives a mean error of $8.98\%$. We note the performance is achieved by training on only a very small amount of data (18 sequences) and without any dataset-specific parameter tuning. We further note that here, without the problems posed by mixed types, the traditional methods are able to reap the benefits of the given geometrical models (an advantage compared to our method which does not have any preconceived model).

%\begin{figure*}[!htb]
%\begin{center}
%\includegraphics[width = 1.00\linewidth]{./Figure/TSNE/TSNE_KT3DMoSeg_v2.pdf}
%\caption{T-SNE visualizations of original feature and learned feature embeddings of KT3DmoSeg sequences. Points are colored by ground-truth}\label{fig:TSNE}
%\end{center}\vspace{-0.5cm}
%\end{figure*}

% Table generated by Excel2LaTeX from sheet ‘AdelaideRMF’
\begin{table}[htbp]
  \centering
  \vspace{-0.2cm}
 \caption{Performance on AdelaideRMF multi-planar (MultiHomo) and motion segmentation (MoSeg). Numbers are in $\%$}\vspace{-0.2cm}
  \footnotesize
   \setlength\tabcolsep{2pt} % default value: 6pt
	  \resizebox{0.72\linewidth}{!}{
   %\setlength\tabcolsep{2pt} % default value: 6pt
    %\resizebox{0.99\linewidth}{!}{
    \begin{tabular}{clllll}
    \toprule
          & \multicolumn{1}{c}{\multirow{2}[3]{*}{\textbf{Model}}} & \multicolumn{2}{p{6em}}{\textbf{MultiHomo}} & \multicolumn{2}{p{6em}}{\textbf{MoSeg }} \\
\cmidrule(lr){3-4} \cmidrule(lr){5-6}          &       & \textbf{Mean} & \textbf{Med.} & \textbf{Mean} & \textbf{Med.} \\
\midrule
    \multirow{5}[1]{*}{\begin{sideways}State-of-the-arts\end{sideways}} & J-Linkage\cite{toldo2008robust} & 25.50 & 24.48 & 16.43 & 14.29 \\
          & T-Linkage\cite{Magri2014} & 24.66 & 23.38 & 9.36  & 7.80 \\
          & RCMSA\cite{pham2014random} & 28.30 & 29.40 & 12.37 & 9.87 \\
          & RPA\cite{Magri2015}   & 17.20 & 17.53 & 5.49  & 4.57 \\
          & ILP-RansaCov\cite{magri2016multiple} & 12.91 & 12.34 & 6.04  & 4.27 \\
    \midrule
    \multirow{4}[4]{*}{\begin{sideways}Ours\end{sideways}} & \textbf{Loss} &       &       &       &  \\
\cmidrule{2-6}          & L2Regress & 17.55 & 14.77 & \textbf{8.98} & 7.50 \\
          & CrossEnt & 17.88 & 12.10 & 9.07  & \textbf{5.79} \\
          & MIMI  & \textbf{17.33} & \textbf{12.00} & 9.39  & 6.50 \\
\cmidrule{2-6}    \end{tabular}%
  \label{tab:AdelaideRMF}%
 % }
 }
 \vspace{-0.5cm}
\end{table}%

\subsection{Further Study}

In this section, we first further analyze the impact of metric learning on transforming the point feature representations. Then we present results on model selection and finally do ablation study for the proposed MaxInterMinIntra loss.

\noindent\textbf{Feature Embedding}
To gain some insight on how the learned feature representations are more clustering-friendly, we provide direct visualization of the representations. For that purpose, we use T-SNE\cite{maaten2008visualizing} to project  both  the KT3DMoSeg raw feature points (of dimension ten for 5 frames) and network output embeddings to a 2-dimensional space. Three example sequences are presented in the last row of Fig.~\ref{fig:KT3DMoSeg}. We conclude from the figure that: (i) the original feature points are hard to be grouped by K-means correctly; and (ii) after our network embedding, feature points are more likely to be grouped according to the respective motions, regardless of the underlying types of motions.
%\begin{figure}[!htb]
%\includegraphics[width=1\linewidth]{./Figure/TSNE/TSNE_Three.pdf}
%\caption{Visualization of point representations before and after metric learning.}\label{fig:TSNE}
%\end{figure}

%We further evaluate the robustness of RANSAC refinement. We notice there are two parameters in this step. First, we determine the rate of over-clustering in K-means. We empirically test the range from $100\%$ to $200\%$. Moreover, we select an outlier threshold for RANSAC and fix for all sequences. We notice the selection of threshold is critical in many conventional multi-model fitting approaches and is often computed from the ground-truth \cite{Magri2014,Magri2015,magri2016multiple}. We present the error rate v.s. threshold under different over-clustering levels in Fig.~\ref{fig:RANSACRefinement}
%
%\begin{figure}
%\includegraphics[width=0.95\linewidth]{./Figure/RANSACRefinement.pdf}
%\caption{RANSAC refinement performance v.s. threshold.} \label{fig:RANSACRefinement}
%\end{figure}

\noindent\textbf{Model Selection}: As can be seen from Fig.~\ref{fig:KT3DMoSeg}, the point distribution in the learned feature embedding is amenable for model selection (estimating the number of clusters/motions). We evaluate both Second Order Difference (SOD)\cite{liu2013robust} and Silhouette Analysis (Silh.) \cite{rousseeuw1987silhouettes} to estimate the number of motions. We also compare with alternative subspace clustering approaches with built-in model selection, namely, LRR \cite{liu2013robust} and MSMC \cite{Dragon2012} and additionally apply self-tuning spectral clustering(S.T.) \cite{zelnik2005self} to the affinity matrix obtained in MVC \cite{xu2018motion}. Performances are evaluated in terms of mean classification error (Err.) and correct rate (Corr.), i.e. the percentage of samples/sequences with correctly estimated number of cluster (higher the better). Comparisons are presented in Tab.~\ref{tab:MdlSel}. Thanks to the deep feature learning, both SOD and Silh. applied to our method give strong performance even though they are very simple heuristics.

% Table generated by Excel2LaTeX from sheet 'ModelSelect'
\begin{table}[htbp]
  \centering
\caption{Comaprison of model selection on KT3DMoSeg. Numbers are in $\%$.}    
\setlength\tabcolsep{4pt} % default value: 6pt
  \resizebox{0.82\linewidth}{!}{
\begin{tabular}{lccccc}
    \toprule
    \multirow{2}[4]{*}{\textbf{Method}} & \multicolumn{2}{c}{\textbf{MIMI Loss}} & \multirow{2}[4]{*}{S.T.\cite{zelnik2005self}} & \multirow{2}[4]{*}{LRR\cite{liu2013robust}} & \multirow{2}[4]{*}{MSMC\cite{Dragon2012}} \\
\cmidrule{2-3}          & SOD\cite{liu2013robust} & Silh.\cite{rousseeuw1987silhouettes} &       &       &  \\
    \midrule
    \textbf{Mean Err $\downarrow$} & 7.36  & \textbf{7.25} & 18.16 & 25.08 & 48.29 \\
    \textbf{Correct $\uparrow$} & \textbf{86.36} & 81.82 & 40.91 & 54.55 & 22.73 \\
    \bottomrule
    \end{tabular}%
    }
  \label{tab:MdlSel}%
\end{table}%

%% Table generated by Excel2LaTeX from sheet 'ModelSelect'
%\begin{table}[htbp]
%  \centering
%  \caption{Comaprison of model selection on LCE and KT3DMoSeg.}
%   \setlength\tabcolsep{4pt} % default value: 6pt
%  \resizebox{0.9\linewidth}{!}{
%    \begin{tabular}{lccccccc}
%    \toprule
%    \textbf{Dataset} & \multicolumn{2}{c}{\textbf{LCE}} & \multicolumn{5}{c}{\textbf{KT3DMoSeg}} \\
%     \cmidrule(lr){2-3} \cmidrule(lr){4-8}
%    \multirow{2}[4]{*}{\textbf{Method}} & \multicolumn{2}{c}{\textbf{MIMI}} & \multicolumn{2}{c}{\textbf{MIMI}} & \multirow{2}[4]{*}{S.T.\cite{zelnik2005self}} & \multirow{2}[4]{*}{LRR\cite{liu2013robust}} & \multirow{2}[4]{*}{MSMC\cite{Dragon2012}} \\
%\cmidrule(lr){2-3}\cmidrule(lr){4-5}          & SOD & Silh. & SOD & Silh. &       &       &  \\
%    \midrule
%    \textbf{Err.} & 18.16 & 19.27 & 7.36  & 7.25  & 18.16 & 25.08 & 48.29 \\
%    \textbf{Corr.} & 97.50 & 76.00 & 86.36 & 81.82 & 40.91 & 54.55 & 22.73 \\
%    \bottomrule
%    \end{tabular}%
%    }
%  \label{tab:MdlSel}%
%\end{table}%

\begin{figure}
\begin{center}
\subfloat{
\includegraphics[width=0.47\linewidth]{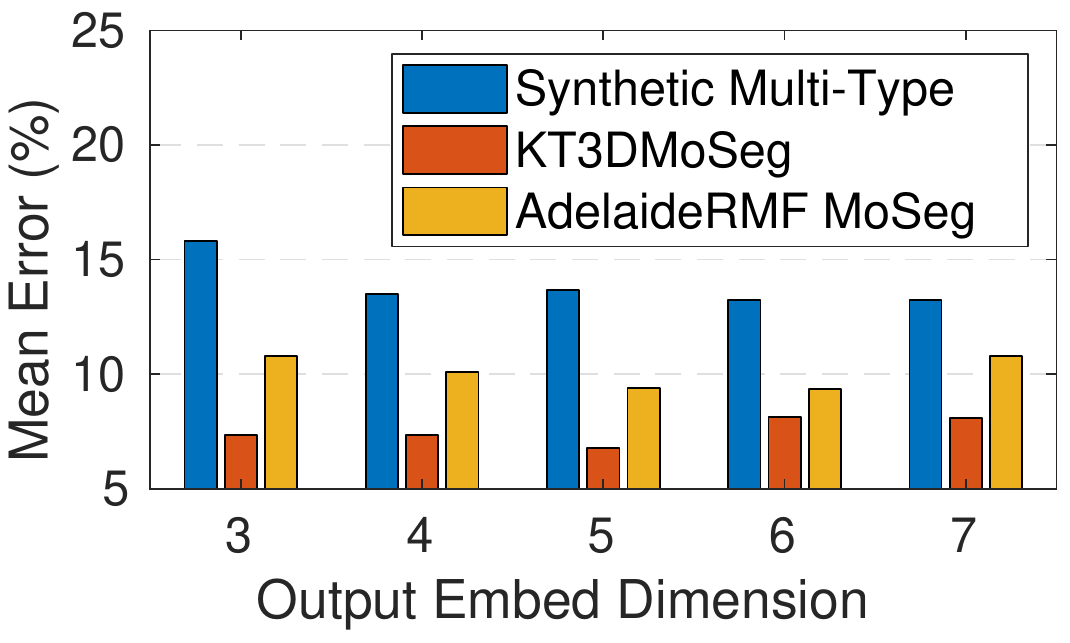}}
\subfloat{
\includegraphics[width=0.38\linewidth]{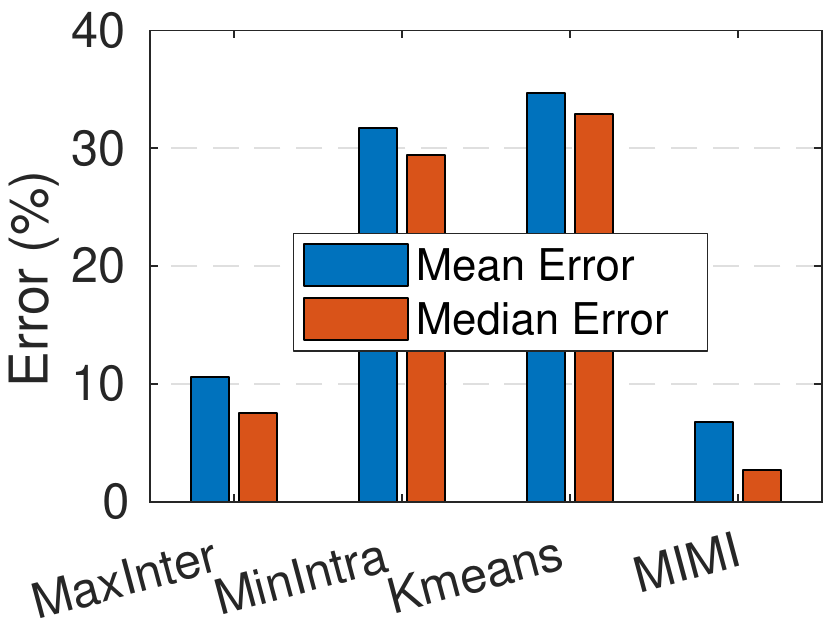}}
\caption{(Left) Performance v.s. the network output dimension. (Right) Comparison of different variants of MIMI loss. }\label{fig:Ablation}
\end{center}
\vspace{-0.6cm}
\end{figure}

\noindent\textbf{Dimension of Output Embedding}:
We investigate the impact of the dimension of the output embedding $\vect{z}$ on the performance of multi-model/type fitting. %We suspect that a lower dimension in the embedding makes the point features less separable, whereas a higher dimension makes the distance metric less effective due to sparsity (don’t understand this?). 
Here, we vary the size of the embedding dimension from 3 to 7 for all three tasks and present the resulting error rates against the dimension in Fig~\ref{fig:Ablation} (left). As we can
see, the errors are relatively stable w.r.t. the output
embed dimension from 4 to 7 for all three tasks with optimal
between 5 to 6 coninciding with the maximal number of
clusters for each task (max 5 motions for KT3DMoSeg and
max 4 structures for Synthetic). Thus
the maximal number of clusters serves as a good heuristic for
the dimension of the network output embedding.

%\begin{figure}
%\begin{center}
%\stackunder[]{
%\includegraphics[width=0.56\linewidth]{./Figure/OutputDim.pdf}}
%\stackunder[]{
%\includegraphics[width=0.44\linewidth]{./Figure/MIMI_KT3DMoSeg_Ablation.pdf}}
%\caption{(a)Performance v.s. the network output dimension.}\label{fig:OutputDim}
%\end{center}
%\end{figure}

\noindent\textbf{MIMI Loss}: %We introduce the MaxInterMinIntra (MIMI) loss for learning good feature for clustering. 
Here we investigate the necessity of both maximizing inter cluster distance and minimizing intra cluster variance. In specific, we compare the following variants. (i) \textbf{MaxInter}: only maximizing the inter cluster distance is considered, equivalent to the first term in Eq~(\ref{eq:MaxInterMinIntra}). (ii) \textbf{MinIntra}: only minimizing the intra cluster variance is considered, the second term in Eq~(\ref{eq:MaxInterMinIntra}). (iii) \textbf{K-means loss}: we further note the k-means loss \cite{yang2017towards} proposed for unsupervised deep clustering shares the same objective with \textbf{MinIntra}. We therefore adapt the k-means loss to supervised learning with fixed point-to-cluster assignment during training. We compare the three variants with our final MIMI loss on KT3DMoSeg and present the results in Fig.~\ref{fig:Ablation} (right). The MIMI loss is consistently better (lower error) than all three variants. In particular, the \textbf{MinIntra} and \textbf{K-means loss} produce large errors. This indicates that pushing points of different clusters away is vital to feature embedding for clustering.

\section{Conclusion}

In this work, we investigate training a deep neural network for general multi-model and multi-type fitting. We formulate the problem as learning non-linear feature embeddings that maximize the distance between points of different clusters and minimize the variance within clusters. For inference, the output features are fed into a K-means to obtain the grouping. Model selection is easily achieved by just analyzing the K-means residual in a parameter free manner. Experiments are carried out on both synthetic and real geometric multi-model multi-type fitting tasks. Comparison with state-of-the-art approaches proves that our network can better deal with multiple types of models simultaneously, without any preconceived notion of the underlying model.  %In fact, this flexibility allows it to learn additional criteria for clustering that is difficult to specify analytically in traditional methods.
 Our method is also less sensitive to sampling imbalance brought about by the increasing number of models, and it works well in a broad range of parameter values, without the kind of careful tuning required in conventional approaches.

%\[
% \begin{matrix}
%  a & b & c \\
%  d & e & f \\
%  g & h & i
% \end{matrix}
%\]

{\small
\bibliographystyle{ieee}
\bibliography{egbib}
}

\end{document}